\documentclass{article}
\pdfoutput=1

\usepackage[preprint]{neurips_data_2023}
\usepackage[utf8]{inputenc} %
\usepackage[T1]{fontenc}    %
\usepackage{hyperref}       %
\usepackage{url}            %
\usepackage{booktabs}       %
\usepackage{amsfonts}       %
\usepackage{amsmath}        %
\usepackage{nicefrac}       %
\usepackage{microtype}      %
\usepackage[dvipsnames]{xcolor}
\usepackage{graphicx}       %
\usepackage{caption}        %

\newcommand{\del}[1]{}  %

\newcommand{\PlantD}{{\textit{Planted}}}  %
\newcommand{\cotwo}{$\text{CO}_2$}  %

\newcommand{\mathdelta}[1]{{\color{blue} ($#1$)}}

\title{\PlantD: a dataset for planted forest identification\\ from multi-satellite time series}

\author{%
Luis Miguel Pazos-Outón\thanks{This work was done while contributing at Google DeepMind.} \\ Mineral AI \And
Cristina Nader Vasconcelos \\ Google DeepMind \And
Anton Raichuk \\ Google DeepMind \And
Anurag Arnab \\ Google DeepMind \And
Dan Morris \\ Google Research \And
Maxim Neumann \\ Google DeepMind
}

\begin{document}

\maketitle

{
    \begin{center}
        \captionsetup{type=figure}
        \begin{minipage}{1\textwidth}
            \centering
            \includegraphics[width=0.9\textwidth]{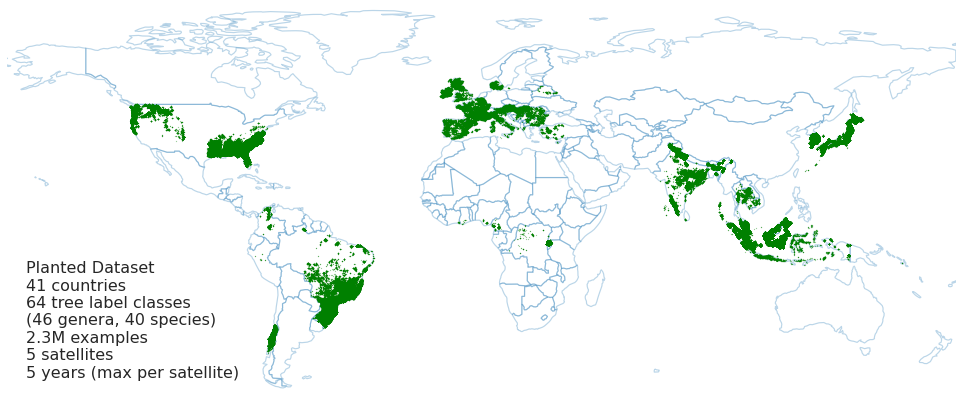}
        \end{minipage}%
        \captionof{figure}{
        Overview of the \PlantD\ dataset and the geographical samples distribution.
        \label{fig:planted-coverage}
        }
    \end{center}
}

\begin{abstract}
Protecting and restoring forest ecosystems is critical for biodiversity conservation and carbon sequestration. Forest monitoring on a global scale is essential for prioritizing and assessing conservation efforts.  Satellite-based remote sensing is the only viable solution for providing global coverage, but to date, large-scale forest monitoring is limited to single modalities and single time points. In this paper, we present a dataset consisting of data from five public satellites for recognizing forest plantations and planted tree species across the globe. Each satellite modality consists of a multi-year time series. The dataset, named \PlantD, includes over 2M examples of 64 tree label classes (46 genera and 40 species), distributed among 41 countries.
This dataset is released to foster research in forest monitoring using multimodal, multi-scale, multi-temporal data sources. Additionally, we present initial baseline results and evaluate modality fusion and data augmentation approaches for this dataset.
\end{abstract}

\section{Introduction}
Forests, covering a third of the Earth's surface, are the most diverse and complex land cover type, and host over 80\% of all terrestrial species. Forests also play a critical role in capturing atmospheric carbon: current estimates suggest that deforestation and land degradation cause 11\% of current carbon emissions, and forest-based solutions can provide 27\% of the mitigation needed to achieve Paris agreement targets, with an annual mitigation potential of 4 gigatons of \cotwo-equivalent emissions per year by 2030~\cite{un2022:glasgow}.

Global monitoring of forests is a promising path to understand the global rates of carbon fluxes and the effectiveness of biodiversity conservation efforts. Remote sensing provides frequent, high-resolution, global imagery, and thus offers a plausible path to global forest monitoring. However, with global coverage comes a significant data review burden that often exceeds the capacity of the conservation community. AI tools are therefore required to fully realize the potential of remote-sensing-based forest understanding. Fortunately, advances in computer vision have been transferred to the remote sensing domain with significant success for detecting forest cover loss and gain \cite{hansen2013:forest-change}, estimating forest canopy height \cite{tolan2023:canopy-height}, and even estimating the carbon content of individual trees \cite{Tucker2023:africa-carbon} at regional to global scales.

At the same time, remote sensing of the Earth's surface from satellite imagery poses its own challenges. Satellite coverage is not uniform around the world, with many temporal, spectral, and geographical data gaps. Different satellite instrument types are sensitive to different forest properties. Data quality and coverage can vary significantly even for individual data sources. Consequently, no single satellite source can provide a complete picture of forest composition, necessitating multi-satellite solutions.

In this work, we construct a planted tree dataset based on a curated subset of the "Spatial Database of Planted Trees" \cite{Harris:sdpt}, extended with several satellite imagery modalities across long time series. Planted forests are regions of planted trees that will be harvested for timber or wood fiber products, and tree crops are regions of planted trees from which products are harvested without removing the trees, for example palm oil, fruit, rubber, and coffee\footnote{In the following we will use the term \textit{forest plantation} or just \textit{plantation} to imply both forest timber plantations and tree crop plantations.}.  As of 2015, about 173 million hectares (4\% of total tree cover) were covered by planted forests, and about 50 million hectares by tree crops \cite{Harris:sdpt}, though the accurate localization of planted forests and tree crops is not possible yet. This dataset offers a challenge for the community to develop approaches to recognize \textit{planted forests} and \textit{tree crops}.

With this work, we aim to foster AI research in remote sensing of forests using multimodal data sources.
Our main contributions are: A curated multimodal dataset for planted forest and tree crop recognition, initial baselines results with a simple multimodal multi-temporal transformer model, and preliminary evaluation of data fusion and data augmentation approaches.

\section{Background}

Remote sensing of forests uses a range of instrument types: optical, multi-spectral, hyper-spectral, synthetic aperture radar (SAR), and lidar \cite{elachi2006:rs-introduction}. These instruments also span a range of spatial resolutions, centimeter scale to kilometer scale. Each instrument and resolution offers a view into different elements of forest composition.

High-resolution optical imagery (30cm to several meters per pixel) has been used, for example, to identify free-standing trees on a regional scale, and to estimate properties related to tree canopy extent, height, and biomass \cite{Tucker2023:africa-carbon, tolan2023:canopy-height}. Multi-spectral imagery, usually providing regular time series at medium resolution (10 to 30 meters per pixel), is frequently used for analyzing forest loss and gain \cite{hansen2013:forest-change} or forest type classification. Hyperspectral imagers provide more and narrower spectral bands, enabling the analysis of fine forest characteristics related to forest health or species composition. Synthetic aperture radar (SAR) and lidar are active sensors, emitting their own electromagnetic waves and measuring the scattered signal from the surface. SAR in particular, due to the use of long wavelengths, is nearly independent of time of day and can penetrate clouds, rain, and parts of the forest canopy. In particular polarimetric and interferometric SAR is sensitive to 3D forest structure and moisture content, and is used for change/anomaly detection, forest type classification, and biomass estimation \cite{papathanassiou2021:forest-polsar}.  Lidar can provide very fine vertical distributions of scattering elements within a small area, and thus can be used to reconstruct the vertical profile of the forest \cite{potapov2021:gedi-forest-mapping}.

\section{Related work}

\subsection{Remote sensing datasets for machine learning}
The application of deep learning to remote sensing data has primarily focused on high-resolution optical imagery, which  has similar characteristics to the natural images that are widely used in computer vision research. Therefore many public remote sensing datasets consist primarily of high-resolution optical imagery (e.g.\ UC Merced \cite{yang2010:ucmerced} or FMoW \cite{christie2018:fmow}). The Sentinel-2 satellite program is the highest-resolution publicly-available optical satellite imagery, with 10m pixel resolution and a revisit time of approximately five days. Consequently, numerous land cover classification datasets have focused on Sentinel-2 data, for example EuroSAT \cite{helber2019:eurosat}, BigEarthNet \cite{sumbul2019:bigearthnet}, or Dynamic World \cite{Brown2022:dynamicworld}. Optical and multi-spectral satellite imagery are often combined with non-optical imagery, such as SAR data from the Sentinel-1 program, for example So2SAT \cite{zhu2018:so2sat}, Sen12 \cite{schmitt2018:sen12}, LandCoverNet \cite{alemohammad2020:landcovernet}, and CropHarvest \cite{tseng2021:cropharvest}. These datasets have broad geographic coverage, but do not provide the granularity of forest categories required to advance forest characterization models.

Forest-specific remote sensing datasets include ReforesTree \cite{Reiersen2022:ReforesTree}, which provides labeled drone imagery over Ecuador; TreeSatAI \cite{Schulz2022:treeSatAI}, which combines Sentinel-1/2 and aerial imagery over a forest region in Germany; and the MultiEarth challenge, which combines imagery from the Sentinel-1, Sentinel-2 and Landsat 8 satellites over the Amazon forest \cite{cha2022:multiearth} with forest segmentation masks. These datasets are specific to forests, but provide narrow geographic coverage and lack category labels that would facilitate the identification of planted forests.

\subsection{Transformers and data fusion}
Transformers, initially introduced for language modeling~\cite{vaswani2017attention}, have been extended to images \cite{dosovitskiy2021:vit}, video~\cite{Arnab2021:vivit}, and audio~\cite{gong2021ast}.
Transformers can naturally be applied to different modalities, since they operate on sequences of abstract tokens. Examples of multimodal fusion with transformers include combining image and text for captioning and visual question answering~\cite{lu2019vilbert, li2019entangled, wang2022ofa}, %
or combining audio and visual information~\cite{nagrani2021attention, georgescu2022audiovisual, jaegle2021perceiver} among others. In the remote sensing domain, Garnot et al.\ \cite{garnot2022:pastis-mm}, for example, evaluated data fusion approaches of Sentinel-1 and Sentinel-2 data for crop type monitoring, where they introduced a variant of the U-Net \cite{Ronneberger2015:unet} model with temporal attention.

\section{The \PlantD\ Dataset}

The \PlantD\ dataset\footnote{The dataset is released under the Creative Commons Attribution 4.0 International (CC-BY-4.0) license, and is available at \url{https://storage.googleapis.com/planted-datasets/public} (version 1.0.1).} is constructed to assess the potential to recognize different types of planted forests and tree crops from diverse satellite data sources. It is a unique, global, multimodal, multi-temporal, multi-scale classification dataset that encourages the development of methods for sensor fusion and time series modeling for forest monitoring applications. Each example contains satellite imagery, labels, and metadata. We create dense imagery cubes for each modality, to cover exactly the same area on the ground at specified date ranges. See \autoref{tab:example-features} for the types and dimensions of all features included in each example.

\subsection{Satellite data}

\begin{table}[ht]
\centering
\caption{List of features in each example in our dataset.}
\label{tab:example-features}
\footnotesize
\begin{tabular}{lllrl}
\toprule
\multicolumn{4}{c}{\bf Satellite data} \\
{\bf \emph{name}} &
{\bf \emph{type}}& 
{\bf \emph{dimensions}} & 
{\bf \emph{description}} \\
\midrule
s1 & float & (8,12,12,3) & Sentinel-1 backscatter \& look angle \\
s2 & int & (8,12,12,10) & Sentinel-2 reflectivity \\
l7 & int & (20,4,4,6) & Landsat-7 reflectivity \\
modis & float & (60,1,1,7) & MODIS reflectivity \& derivatives \\
alos & float & (4,4,4,3) & ALOS PALSAR backscatter \& look angle\\
s1\_timestamps & int & (8) & Sentinel-1 timestamps \\
s2\_timestamps & int & (8) & Sentinel-2 timestamps \\
l7\_timestamps & int & (20) & Landsat 7 timestamps \\
modis\_timestamps & int & (60) & MODIS timestamps \\
alos\_timestamps & int & (4) & ALOS PALSAR timestamps\\
s1\_mask & int & (8,12,12,3) & Sentinel-1 mask \\
s2\_mask & int & (8,12,12,10) & Sentinel-2 mask \\
l7\_mask & int & (20,4,4,6) & Landsat 7 mask \\
modis\_mask & int & (60,1,1,7) & MODIS mask \\
alos\_mask & int & (4,4,4,3) & ALOS PALSAR mask\\
\midrule
\multicolumn{4}{c}{\bf Labels}
\\
{\bf \emph{name}} & 
{\bf \emph{type}}& 
{\bf \emph{}}&  
{\bf \emph{description}}\\
\midrule
common\_name & string & & Common name of the tree species\\
species & string & & Species name\\
conifer\_broad & string & & Whether conifer or broad-leaf\\
ever\_dec & string & & Whether evergreen or deciduous\\
hard\_soft & string & & Whether hard or soft wood\\
\midrule
\multicolumn{4}{c}{\bf Metadata}
\\
{\bf \emph{name}} & 
{\bf \emph{type}}& 
{\bf \emph{unit}}&  
{\bf \emph{description}}\\
\midrule
country & string & & Country name\\
area\_ha & float & ha & Area of forest plot, in hectares\\
perimeter\_km & float & km & Perimeter of forest plot, in kilometers\\
elevation & float & m & Elevation above sea surface, in meters\\
lat & float & deg & Latitude, in degrees\\
lon & float & deg & Longitude, in degrees\\
\bottomrule
\end{tabular}
\end{table}

\begin{table}[ht]
\centering
\caption{Satellite data characteristics. Columns: Satellite name, abbreviation, instrument type, resolution in meters, frequency of data collection, time range for temporal aggregation, start of time series, tensor dimensions in each example (T,H,W,C) = (time series samples, height, width, number of channels). }
\label{tab:sats}
\footnotesize
\scalebox{.95}{
\begin{tabular}{lccccccc}
\toprule
{\bf Satellite} & {\bf Instrument} & {\bf Res.} & {\bf Frequency} & {\bf Temporal agg.} & {\bf Time range} & {\bf Start} & {\bf Dimensions} \\
\midrule
Sentinel-1 & SAR C-band & 10m & 5 days & seasonal & 2 years & 01/2016 &(8,12,12,3) \\
Sentinel-2 & Multi-spectral & 10m & 5 days & seasonal & 2 years & 01/2016 &(8,12,12,10) \\
Landsat 7 & Multi-spectral & 30m & 16 days & seasonal & 5 years & 01/2013 & (20,4,4,6) \\
ALOS-2 & SAR L-band & 30m & 6-30 days & yearly & 3 years & 01/2015 & (4,4,4,3) \\
MODIS & Multi-spectral & 250m & daily & monthly & 5 years & 01/2013 & (60,1,1,7) \\
\bottomrule
\end{tabular}
}
\end{table}

Each data example contains image patches across time for the five satellite data sources covering an area of 120 meters x 120 meters. Due to differing resolutions among the satellites, the actual image sizes (height and width) vary from satellite to satellite. Additionally, each satellite modality has a different number of time series samples and bands.
Table \ref{tab:sats} provides an overview of the sample dimensions, pixel resolution, and temporal characteristics per satellite.

Satellite data for each example contains a multi-year time series. The start and end times of the time series within an example are not the same for all satellites due to different start times and duration of data acquisition. To reduce the data size, we aggregate images based on satellite-specific temporal reduction functions. Temporal reductions can be monthly, seasonal (i.e. 3 months), or yearly. MODIS and Landsat 7 have full data coverage over the time range (2013-2017), while others do not due to later satellite launches and data availability. Some examples have missing images due to having missing or low-quality filtered-out data. This is a common issue when dealing with remote sensing data, either due to meteorological, geographical, or instrument-specific constraints. The satellite-dependent temporal distribution of invalid samples is presented in the supplementary material.

For each satellite image source and each time sample we added timestamps (mean of the time range) as an integer, counting the number of milliseconds since \textit{1970-01-01T00:00:00Z}. For convenience, we included date strings in the format \textit{YYYY-MM-DD}. Finally, each image patch is accompanied by a binary mask of pixel validity.

The following imaging instruments are included in the \PlantD\ dataset:
\begin{itemize}
\item Sentinel-1 (S1): Synthetic aperture radar (SAR) providing data since 2014.
\item Sentinel-2 (S2): Multispectral satellite providing data since 2015.
\item Landsat 7 (L7): Multispectral satellite imagery, provided data from 1999 to 2022.
\item ALOS-2 (Advanced Land Observing Satellite) PALSAR-2: Synthetic aperture radar (SAR), providing data since 2014.
\item MODIS (Moderate Resolution Imaging Spectroradiometer): Medium resolution multispectral satellite imagery, providing data since 1999.
\end{itemize}
For more details about satellite characteristics and individual bands, see the supplementary material.

\subsection{Labels}
The \PlantD\ dataset consists of 64 forest and plantation tree classes, from 41 countries, and a total of 2,264,747 examples. The geographic distribution of samples is shown in \autoref{fig:planted-coverage}. Each example includes the common name class (64), genus (46), and, when known, the tree species (40). The labels are based on a subset of \cite{Harris:sdpt}, having been curated as described in \autoref{sec:dataset-gen}. The approximate years of data labeling are 2013-2015, though the exact time associated with each label is not known.  In addition to the tree species, the categorization into broad-leaf vs.\ conifer, evergreen vs.\ deciduous, hard- vs.\ soft-wood, and planted trees vs.\ tree crops are provided.

The dataset is unbalanced, with numbers of examples for a given class ranging between 15 and 450k. In order to facilitate more reliable evaluation, we introduce a frequency annotation to each species label. Categories with less than 200 samples are assigned to the \textit{rare} sub-split, categories with $\geq$ 200 and $<$ 10k samples are assigned to the \textit{common} sub-split, and those with $\geq$ 10k are assigned to the \textit{frequent} sub-split. All species and their frequency annotations are listed in the Appendix.

\subsection{Metadata}
Additional metadata about the forest plantations includes the country, the area of the plantation in hectares, the perimeter of the plantation, the elevation in meters, and the center latitude and longitude in degrees.

\subsection{Examples analysis}
\begin{figure}
    \centering
    \includegraphics[width=\linewidth]{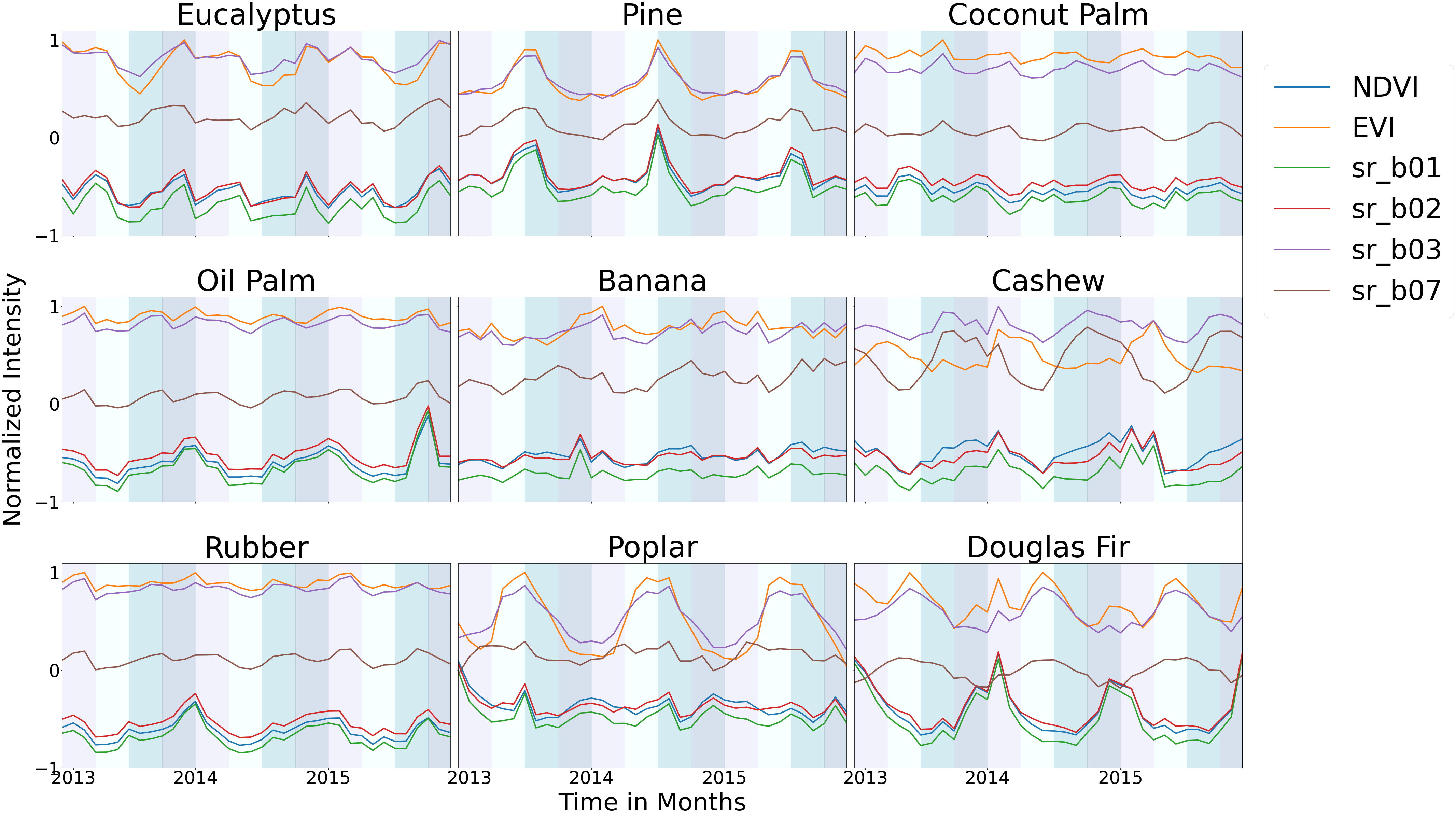}
    \caption{Temporal signatures of randomly selected tree species across MODIS spectral reflectance bands and NDVI and EVI indicators.}
    \label{fig:spectral_1}
\end{figure}

The temporal signatures of MODIS spectral bands and derived indicators for ten randomly selected tree species are shown in Figure \ref{fig:spectral_1}. The date range covers the years from 2013 to 2016. Each band is normalized to have a mean of 0 and a standard deviation of 1. It illustrates and emphasizes the importance of temporal and spectral signatures for forest monitoring -- the signatures differ significantly among tree species and correlate with seasonality.

\section{Methods}

\subsection{Dataset generation}
\label{sec:dataset-gen}

The labels are based on the Spatial Database of Planted Trees (v1.0) \cite{Harris:sdpt}.
We performed an extensive analysis of the provided data and selected a subset of the examples for the first version of the \PlantD\ dataset based on the following criteria:
\begin{enumerate}
\item We explicitly filtered out examples which have been labeled by another machine learning approach --- in order not to distill a potentially weaker model. The source labels should be based on manual labeling or surveys.
\item The species information should be known exactly. We keep only monoculture forest plantations and drop mixtures or poorly defined species examples.
\item The minimum area of forest plantation should be at least one quarter of a hectare.
\item We de-duplicated samples locations to be at least 70m from each other.
\item After those steps, the final filter included dropping species with less than 10 examples.
\end{enumerate}
This resulted in a global dataset, consisting of 64 forest and plantation tree species, from 41 countries, and a total of 2,264,747 examples.

For satellite data processing and extraction of the localized regions of interest (ROIs), we used Google Earth Engine \cite{gorelick:gee:2017}. The following processing steps were performed:
\begin{enumerate}
\item Constructing deterministic data ranges for individual satellite data (yearly for ALOS, monthly for MODIS, and seasonally for Sentinel-1/2 and Landsat 7).
\item Filtering out bad data (e.g. based on cloud cover).
\item Spatial re-sampling of all bands at satellite-dependent nominal resolution.
\item Adding a mask for invalid pixels or invalid images.
\item Aggregation of multiple samples within a given date range into a single multi-band image: The reduce function for Sentinel-1, ALOS, MODIS was the \textit{mean}, while for Sentinel-2 and Landsat 7 it was a \textit{cloud-coverage-based mosaic}.
\item Extraction of resolution-dependent multi-band images with a size of 120 meters x 120 meters centered at the forest plantation center.
\end{enumerate}

The examples were distributed into three splits (\textit{train, validation, and test}), with a reference ratio of 8:1:1, respectively. However, to ensure reliable evaluation, for classes with less than 100 total examples, we used a 1:1:1 ratio, and we limited the number of validation and test examples to about 1000. We used an adaptive \textit{geographic splitting} approach, first dividing the world into regions of approximately 20x20 km (400 km${}^2$) to 80x80 km (6400 km${}^2$) (depending on how many such regions we got per label), and assigning all samples within one such region randomly to one of the three splits. The distribution of labels across the splits is available in the Appendix.

\subsection{Transformer model}

\begin{figure}
    \centering
    \includegraphics[width=0.9\linewidth]{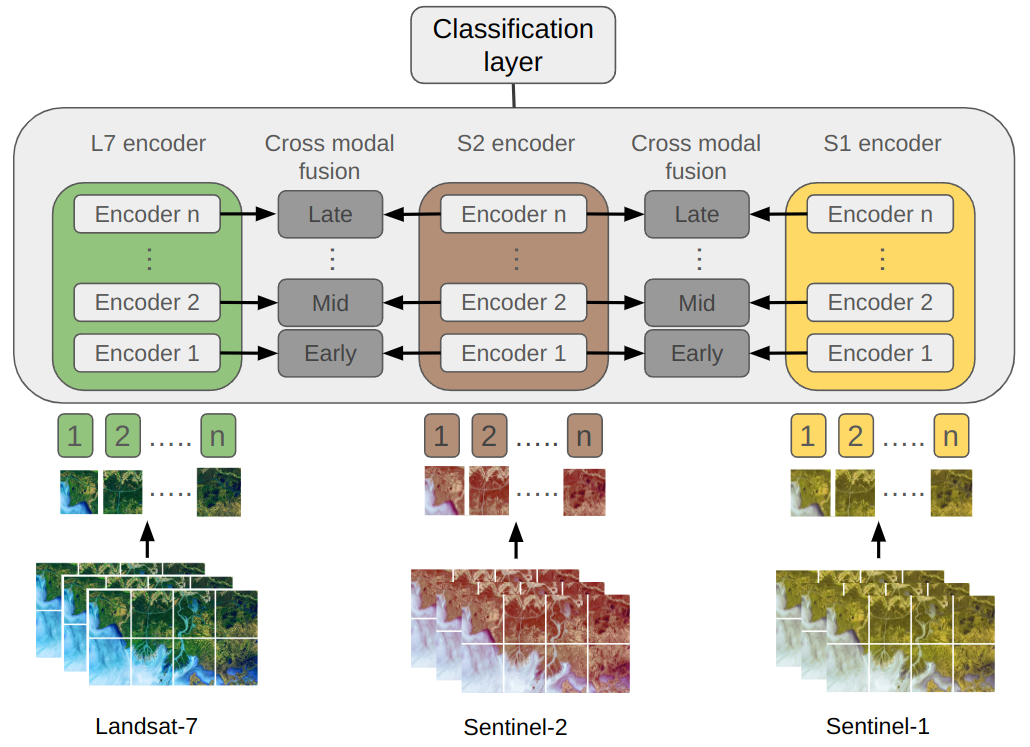}
    \caption{Depiction of the multimodal transformer architecture. In early fusion all cross-modal layers are activated, in mid fusion the early layers are excluded and we rely on the mid layer and the final layers. In late fusion only the last cross-modal layers are used.}
    \label{fig:transformer_illustration}
\end{figure}

Transformers are generic architectures based on non-local attention mechanism that operate on any sequence of tokens. Vision transformers~\cite{dosovitskiy2021:vit} obtain tokens by linearly projecting non-overlapping, 2D patches from the input image. Transformer models for video~\cite{Arnab2021:vivit} extend the tokenization process to 3D patches (or \textit{tubelets}) in order to include the temporal axis, as conceptually visualized in \autoref{fig:transformer_illustration}.

In this dataset, most satellites contain both spatial and temporal dimensions, and we therefore tokenize them with 3D patches. The exception to this is MODIS, for which each pixel is larger than the patch size, so MODIS can be seen as a 1D time series which is tokenized with a 1D patch.

\subsection{Multimodal fusion}
Straightforward approaches for fusing multiple modalities include \textit{early fusion} and \textit{late fusion}.
In \textit{early fusion}, we extract tokens from each modality, concatenate them, and process them with a transformer.
\textit{Late fusion}, in contrast, processes each modality with a separate model. Tokens from each modality are then concatenated together before being passed to a linear classifier.

\textit{Early fusion}~\cite{karpathy2014large} is simple, but computationally expensive, as the sequence length increases linearly with the number of modalities, and computation in transformers has quadratic complexity with respect to the number of tokens.

\textit{Late fusion} is practical, as it allows us to re-use models already trained for a single modality~\cite{karpathy2014large,simonyan2014two}.
The number of model parameters, however, grows linearly with the number of modalities.

\textit{Mid fusion}~\cite{nagrani2021attention,garnot2022:pastis-mm} strikes a balance between \textit{early} and \textit{late} fusion, by initially processing modalities independently of one another, before fusing them at an intermediate point in the network.

\section{Experimental results}

\subsection{Experimental setup}

For the initial baseline experiments we used a standard vision transformer \cite{dosovitskiy2021:vit} architecture extended to multi-temporal 3D data inputs as in video \cite{Arnab2021:vivit}. Our default baseline transformer encoder architecture contains 12 layers with three self-attention heads, with an embedding size of 192 and an MLP size of 768. Pre-processing consists of per-modality input data normalization using robust statistics in order to minimize the effect of outliers, which are common in remote sensing data (median for center and median absolute deviation (MAD) for scale). Optionally, data augmentations during training are performed as described in the experiments below. By default, we use the Adam optimizer with a learning rate of 0.001 and weight decay of 0.0001, minimizing the softmax cross-entropy loss.

We evaluate the performance using the \textit{overall accuracy} and the \textit{macro F1 score} metrics, which we report for the entire split. We further break down the F1 metrics into \textit{rare}, \textit{common}, and \textit{frequent} sub-splits.
The overall accuracy is micro-averaged, considering each example independently. The F1 score is macro-averaged across the classes. It is defined as:
\begin{equation}
    \text{F1} = \frac{1}{C} \sum_c^C \frac{2 TP_c}{2 TP_c+FP_c+FN_c}
\end{equation}
where $C$ is the number of classes, and $TP_c$, $FP_c$, $FN_c$ are the true positives, false positives, and false negatives of class $c$.
For reporting F1 on label frequency sub-splits, we average per-class F1${}_c$ over the set of labels in the \textit{rare}, \textit{common}, and \textit{frequent} categories.

\subsection{Embedding patch sizes}
\begin{figure}
    \centering
    \includegraphics[width=0.75\linewidth]{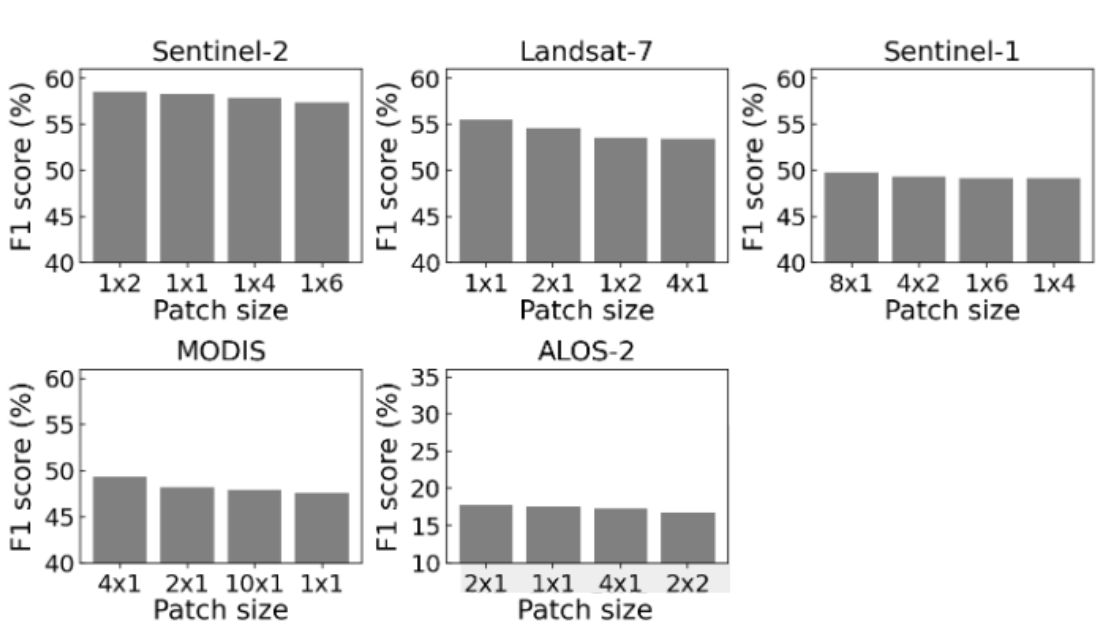}
    \caption{Patch size ablations for each satellite. The best  patch size is satellite-dependent. The notation above refers to time x spatial dimensions.}
    \label{fig:patch_size_ablation}
\end{figure}

In the first series of experiments, we explore performance of basic transformers using only a single satellite data modality. Since the spatio-temporal image sizes vary significantly among satellites, we first investigate different patching strategies.
Figure \ref{fig:patch_size_ablation} shows the performance of different patch sizes.
A patch size, denoted as $T\times S$, implies a 3D patch of temporal size $T$, and spatial height and width sizes $S$.
As we can see, for the current model configuration and processing, the optimal patch size for MODIS is 4x1, for Landsat 7 is 1x1, for Sentinel-1 is 8x1, for Sentinel-2 is 1x2, and for ALOS is 2x1. That is, we can observe that spatial patching leans toward higher  resolution (i.e. small or no patching), while temporal aggregation within tokens is sometimes preferred (Sentinel-1, MODIS).

\subsection{Training data augmentation}

Many image augmentation techniques in computer vision adopt image transformations that are based on human perception invariances \cite{Cubuk_2019_CVPR}. However, in remote sensing, texture often carries high information content, despite being difficult for humans to perceive. Therefore initially we limit our augmentations to affine spatial transformations (rotation and flip) and temporal masking (similar to temporal dropout in \cite{garnot2022:pastis-mm}).
We explore three different data augmentation policies: (i) training with no data augmentation;
(ii) training with rotation and flip; and (iii) temporal masking.

In \autoref{tab:data_augm_single_sat} we show the performance of different satellites with and without augmentations. Spatial augmentation usually helped, and we observed improvements of $1\%$ to $3\%$ with temporal masking, except for Sentinel-2 and MODIS. MODIS has the longest time series with high redundancy, which is why we hypothesize the temporal augmentation was not helpful. Sentinel-2 on the other hand has the highest amount of missing data. These results were surprising and we are continuing to investigate it.

\begin{table}[bt]
\caption{
Impact on {\bf Accuracy} of data augmentation on training individual satellites. Columns show performance (i) with no data augmentation; (ii) adopting rotation and flip, and (iii) adding temporal masking with probability set as $50\%$.}
\label{tab:data_augm_single_sat}
\centering
\begin{tabular}{lcccrr}
\toprule
\multicolumn{1}{c}{\bf Satellites} &
\multicolumn{3}{c}{\bf Data Augmentation} \\ 
&
{\bf \emph{none}} & 
{\bf \emph{rotation \& flip}} & {\bf \emph{temporal masking}} 
\\ 
  \cmidrule(r){2-4}
Sentinel-1 & {51.0} & {50.66}  \mathdelta{-0.34} & {52.41}  \mathdelta{1.41} \\
Sentinel-2 & {56.37} & {56.55}  \mathdelta{0.17} & {60.13}  \mathdelta{3.75} \\
Landsat-7 & {54.25} & {52.56}  \mathdelta{-1.69} & {56.26}  \mathdelta{2.01} \\
MODIS & {45.23} & {44.55}  \mathdelta{-0.67} & {48.34}  \mathdelta{3.12} \\
ALOS-2 & {28.63} & {23.13}  \mathdelta{-5.5} & {28.62}  \mathdelta{-0.01} \\
\bottomrule\\
\end{tabular}
\end{table}

\subsection{Fusing multiple modalities}
In order to maximize the benefits of all modalities available, we explored different approaches for multimodal fusion. The optimal patch size derived for each individual satellite in the previous section was used here when combining multiple satellites.

\autoref{tab:best_performing} presents the best selected configuration as we increase the number of satellites. The selection was done based on the validation F1 macro score. We also present the sub-split and test metrics. Interestingly, the best validation accuracy and F1 scores are obtained using 2 or 3 satellite data sources. As expected, we observe significant differences between rare, common, and frequent classes, ranging from 26.1\% F1${}_r$ to 87.3\% F1${}_f$, with F1${}_c$ falling between the two. An important challenge for future research could be to improve F1${}_r$ and F1${}_c$.

We also evaluated the performance of non-transformer models commonly used in remote sensing; namely a fully connected network and a random forest. Those models were optimized using AutoML on the cases of a single satellite and all satellites. The best test F1 Macro was achieved by the fully connected network, with a value of 51.8\%. The random forest showed a best value of 41.6\%.

\begin{table}
\centering
\caption{Validation, Test Accuracy and F1 score for best performing groups of satellites and sub-splits ({\bf F1${}_r$},  
{\bf F1${}_c$} and {\bf F1${}_f$} denote
\textit{rare}, \textit{common} and \textit{frequent} classes
respectively). Selection of best models based on validation accuracy. }
\label{tab:best_performing}
\resizebox{\textwidth}{!}{
\large
\begin{tabular}{lrrrrr@{\hskip 0.21in}rrrrr}
\toprule
{\bf Satellites} &
 \multicolumn{5}{c}{\bf Validation}& \multicolumn{5}{c}{\bf Test}
\\
& {\bf Acc.}&{\bf F1 Macro}&{\bf F1${}_r$}&{\bf F1${}_c$}&{\bf F1${}_f$} 
& {\bf Acc.}&{\bf F1 Macro}&{\bf F1${}_r$}&{\bf F1${}_c$}&{\bf F1${}_f$} 
\\\cmidrule(r){2-6}
  \cmidrule(r){7-11}
s2&\textbf{96.1}&60.3&26.1&51.3&86.9&96.1&61.7&31.7&50.4&87.7\\
s1-s2&95.7&\textbf{61.4}&31.2&50.3&\textbf{87.3}&95.8&\textbf{62.2}&\textbf{34.8}&50.1&87.3\\
alos-modis-s2&96.0&\textbf{61.4}&30.4&\textbf{51.8}&86.7&\textbf{96.2}&62.1&30.0&\textbf{52.1}&\textbf{88.3}\\
alos-modis-s1-s2&95.6&61.1&\textbf{32.5}&50.0&86.2&95.7&61.2&31.7&50.6&86.3\\
l7-alos-modis-s1-s2&95.0&59.9&\textbf{32.5}&47.4&85.1&95.2&59.3&30.7&47.1&85.1\\
\bottomrule
\end{tabular}
}
\end{table}

\begin{figure}[ht!]
    \centering
    \includegraphics[width=1.0\linewidth]{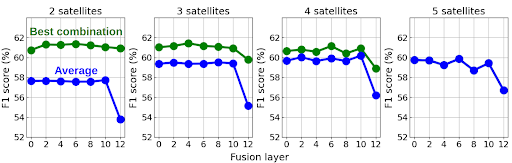}
    \caption{Performance for different fusion layers. The green line shows the top performing combination of satellites, and the blue line the average performance across all combinations of satellites.}
    \label{fig:where_to_fuse_mean}
\end{figure}

\subsection{Early vs.\ late fusion}

To identify a good generic layer at which to fuse modalities, we evaluated the performance of all satellite combinations when fusing modalities at different transformer layers. After averaging across all modality combinations, \autoref{fig:where_to_fuse_mean} illustrates the performance of the best satellite combination and the average across all combinations.

We observe that optimal performance is typically achieved using mid fusion. The worst performance is always observed when fusing in the last layer, suggesting that the simple fusion of independent models is not an efficient approach to combine multiple satellites. Instead, the different transformer backbones should be allowed to exchange information earlier in the stack to maximize the distillation of satellite information.

\section{Conclusion}
We present the \PlantD\ dataset for global planted forest and tree crop recognition from multi-satellite time series. This dataset addresses a critical need in the sustainability space (the delineation of planted forests), and further offers AI researchers a diverse and challenging problem.

This dataset poses challenging questions for exploration: How can we efficiently fuse information from multiple satellite sources? What are the most effective pretraining strategies for multimodal remote sensing tasks? What are the optimal methods for handling dataset imbalances?

We present initial baseline results evaluating single-modality time-series models as well as multimodal fusion. We hope these results can serve as a benchmark for the community, and that users of this dataset can build upon our methods to develop models that transfer well across geographies, satellite modalities and time. We hope this helps the global forestry community to develop comprehensive, accurate, planet-scale forest monitoring.

\section*{Acknowledgments}
We would like to thank Nancy Harris, Liz Goldman and Sarah Carter from the World Resources Institute for advice regarding the SDPT data. We thank Ross Goroshin for designing initial figures for MODIS temporal signatures and Vishal Batchu for advising us on Google Earth Engine pipelines.

Data sources acknowledgments:
Landsat 7 image courtesy of the U.S. Geological Survey.
MODIS data courtesy of the U.S. Geological Survey \cite{modis}.
Sentinel-1 and Sentinel-2 data courtesy of the ESA Copernicus program. Contains modified Copernicus
Sentinel data 2015-2017.
 ALOS-2 PALSAR-2 data courtesy of JAXA (Japan Aerospace Exploration Agency).
Aggregated and harmonized forest plantations labels data courtesy of Nancy Harris, Elizabeth Goldman and Samantha Gibbes, "Spatial Database of Planted Trees (SDPT Version 1.0)", World Resources Institute, 2019 \cite{Harris:sdpt}.

\bibliographystyle{authordate1}
\bibliography{main}

\clearpage
\appendix

\section{Satellite data details}
\label{app:details-sats}
This section describes the used satellite data in more detail and includes tables of data bands included in the \PlantD\ dataset.

\paragraph{Sentinel-1 (S1)} satellites carry a Synthetic Aperture Radar (SAR) at C-band frequency (center wavelength of 5.54 cm (5.4 GHz), with a bandwidth of 100 MHz). It images land and ice surfaces of the Earth between -79 and +82 degrees latitude. It can send and receive at vertical and horizontal polarizations, though we include data only for the more common VV and VH combinations (vertical transmit, and vertical and horizontal receive). Currently there is only a single satellite in operation (Sentinel-1A, launched in 2014) with a revisit time on each point on Earth of 12 days. Sentinel-1B, launched in 2016 had a malfunction and provided data only till Dec 2021 (in this time range the revisit time for both satellites was 6 days). Sentinel-1C is expected to be launched in 2023.
We extract the backscatter at VV and HV polarizations, which are already converted to decibel (dB). Additional pre-processing within Google Earth Engine includes thermal noise removal, radiometric calibration, and terrain correction. SAR data is very sensitive to the incidence angle at the surface, and therefore we also include the used incidence angle with respect to reference Earth ellipsoid (\autoref{tab:bands-s1}).

\begin{table}[htbp]
\centering
\caption{Included bands from Sentinel-1.}
\label{tab:bands-s1}
\begin{tabular}{lll}
\toprule
Index & Bands & Description \\
\midrule
0 & VV &  Vertically polarized transmit and receive, in dB \\
1 & VH & Vertically polarized transmit and horizontally polarized receive, in dB \\
2 & angle & Incidence angle  \\
\bottomrule
\end{tabular}
\end{table}

\paragraph{Sentinel-2 (S2)} The satellite constellation of ESA's Copernicus program contains of 2 satellites with medium-resolution multi-spectral imagers (as of 2023): Sentinel-2A was launched in June 2015 and Sentinel-2B in March 2017 (Sentinel-2C expected to launch in 2024, and Sentinel-2D in 2025). Each satellite revisits the same point on Earth every 10 days. Since they operate at an offset, they image the target Earth regions (land and coastal, between -56 and 83 degrees latitude) every 5 days at the equator (and with higher frequency towards the poles). Sentinel-2's are polar-orbiting sun-synchronous (always imaging the Earth at 10:30 local time) with a ground swath of 290 km.
S2 is a multi-spectral instrument, measuring reflectance of sun's illumination from the Earth in 13 bands with different center frequencies (between 0.442 $\mu$m and 2.186 $\mu$m, bandwidths (0.02 $\mu$m to 0.185 $\mu$m) and pixel resolutions (10 m to 60 m). The list of bands used in \PlantD\ dataset is presented in \autoref{tab:bands-s2}. The bands B1, B9 and B10 are used mostly for clouds/atmosphere. B2-B4 are blue, green, red bands. B5-B8A are vegetation red edge and near-infra-red (NIR) bands sensitive to fine vegetation characteristics. Finally, longer wavelength bands B11-B12 are short-wave infrared (SWIR) bands with sensitivity to moisture and fine structural properties.

\begin{table}[htbp]
\centering
\caption{Included bands from Sentinel-2 (reference numbers from Sentinel-2A).}
\label{tab:bands-s2}
\begin{tabular}{llrrrl}
\toprule
Index & Band & Wavelength & Bandwidth & Resolution & Description \\
\midrule
0 & B2 & 492.4 nm & 66 nm & 10 m & Blue \\
1 & B3 & 559.8 nm & 36 nm & 10 m & Green \\
2 & B4 & 664.6 nm & 31 nm & 10 m & Red \\
3 & B5 & 704.1 nm & 15 nm & 20 m & Vegetation red edge 1 \\
4 & B6 & 740.5 nm & 15 nm & 20 m & Vegetation red edge 2 \\
5 & B7 & 782.8 nm & 20 nm & 20 m & Vegetation red edge 3 \\
6 & B8 & 832.8  nm & 106 nm & 10 m & Near infrared (NIR) \\
7 & B8A & 864.7 nm & 21 nm & 20 m & Narrow NIR \\
8 & B11 & 1612.7 nm & 91 nm & 20 m & Short wave infrared (SWIR) 1 \\
9 & B12 & 2202.4 nm & 175 nm & 20 m & Short wave infrared (SWIR) 2 \\
\bottomrule
\end{tabular}
\end{table}

\paragraph{Landsat-7 (L7)} the seventh satellite of NOAA and NASA's Landsat program, was launched in April 1999 and was operational far beyond the envisioned five year mission duration, and was decommissioned only in 2021. Similar to Sentinel-2 it has a polar, sun-synchronous orbit, with equatorial crossing time of about 10:00 am, and with a 16 days revisit time needed to scan the entire Earth. The main imaging instrument on board of Landsat 7 is the Enhanced Thematic Mapper Plus (ETM+), providing 8 bands. For the \PlantD\ dataset we keep only the 30m resolution bands (\autoref{tab:bands-l7}, dropping the panchromatic (B8) and the thermal (B6) bands.

\begin{table}[htbp]
\centering
\caption{Included bands from Landsat-7.}
\label{tab:bands-l7}
\begin{tabular}{llrrrl}
\toprule
Index & Band & Wavelength & Bandwidth & Resolution & Description \\
\midrule
0 & B1 & 485 nm & 70 nm & 30 m & Blue \\
1 & B2 & 560 nm & 80 nm & 30 m & Green \\
2 & B3 & 660 nm & 60 nm & 30 m & Red \\
3 & B4 & 835 nm & 130 nm & 30 m & Near infrared (NIR) \\
4 & B5 & 1650 nm & 200 nm & 30 m & Short wave infrared (SWIR) 1 \\
5 & B7 & 2215 nm & 270 nm & 30 m & Short wave infrared (SWIR) 2 \\
\bottomrule
\end{tabular}
\end{table}

\paragraph{ALOS-2} carries the PALSAR-2 SAR instrument at L-band frequency (center wavelength: 22.9 cm, 1.2 GHz; bandwidth: 14-84 MHz). It is the third satellite with an L-band SAR from the Japan Aerospace Exploration Agency (JAXA): first was JERS-1 launched in 1992, then ALOS/PALSAR launched in 2006, while ALOS-2/PALSAR-2 was launched in 2014. Due to larger wavelength than C-band from Sentinel-1, ALOS electromagnetic waves can penetrate deeper into the forest canopy and are more sensitive to tree trunks and branches, and can get a significant ground-trunk double-bounch scattering contribution. Similar to Sentinel-1, it is a fully polarimetric SAR sensor, though the majority of data is available in HH and HV combination. We use the pre-processed yearly mosaics as they are provided by Google Earth Engine. The HH and HV backscatter magnitudes (\autoref{tab:bands-alos} are represented by 16-bit digital numbers (DN) and they can be converted to backscatter values in decibels (dB) via:
\[
    \gamma = 10\log_{10}(\text{DN}^2) - 83.0\ \text{dB}
\]

\begin{table}[htbp]
\centering
\caption{Included bands from ALOS-2 PALSAR-2.}
\label{tab:bands-alos}
\begin{tabular}{lll}
\toprule
Index & Bands & Description \\
\midrule
0 & HH &  Horizontally polarized transmit and receive \\
1 & HV & Horizontally polarized transmit and vertically polarized receive \\
2 & angle & Incidence angle  \\
\bottomrule
\end{tabular}
\end{table}

\paragraph{MODIS} (Moderate Resolution Imaging Spectrometer) is a multi-spectral imaging instrument on board of Terra and Aqua satellites. It has much lower spatial resolution (250m to 1km) than Sentinel-2 or Landsat-7, but it images the Earth (land, ocean, and atmosphere) in more frequency bands (36), and at more frequent intervals (every 1-2 days). Terra was launched in December 1999, and Aqua was launched in May 2002. Both fly on near-polar, sun-synchronous, circular orbits. Of importance for forest monitoring are especially the first seven bands, while others are used for monitoring the ocean and the atmosphere.

\begin{table}[htbp]
\centering
\caption{Included bands from MODIS Terra/Agua satellites.}
\label{tab:bands-modis}
\begin{tabular}{llrrrl}
\toprule
Index & Band & Wavelength & Bandwidth & Resolution & Description \\
\midrule
0 & B01 & 645 nm & 50 nm & 250 m & Red \\
1 & B02 & 858 nm & 35 nm & 250 m & Near infrared (NIR) \\
2 & B03 & 469 nm & 20 nm & 500 m & Blue \\
3 & B04 & 555 nm & 20 nm & 500 m & Green \\
4 & B05 & 1240 nm & 20 nm & 500 m & Short wave infrared (SWIR) \\
5 & B06 & 1640 nm & 24 nm & 500 m & Short wave infrared (SWIR) \\
6 & B07 & 2130 nm & 50 nm & 500 m & Short wave infrared (SWIR) \\
\bottomrule
\end{tabular}
\end{table}

\subsection{Missing data and samples availability per satellite}
\label{app:missing-data}
\begin{figure}[htbp]
    \centering
    \includegraphics[width=0.5\linewidth]{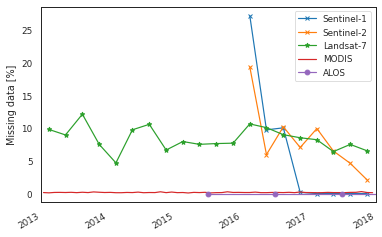}
    \caption{Percentages of missing data over time per satellite (based on 200,000 random examples from the train split).}
    \label{fig:missing-data}
\end{figure}

Despite temporal aggregation of satellite observations often some data still remains missing. \autoref{fig:missing-data} outlines the percentages of missing data across time. ALOS data, partially due to yearly aggregation is nearly always available to 100\%. MODIS data is usually available (99.8\%). Landsat-7 data is missing in about 8.6\% of data, as does Sentinel-2. Sentinel-1 misses 6.9\%. Since Sentinel-1 and -2 were launched the latest, it takes time till the data becomes stable and well calibrated. Therefore we see initially large amounts of missing data in early 2016 (31.4\% for S1 and 21.7\% for S2) which falls over the 2 year range to 0\% for S1 and 2.5\% for S2.

\section{Labels data details}
\label{app:details-labels}

\autoref{tab:labels-distribution} presents the distribution of species across train, validation and test splits. It also denotes ratio of train examples to the total number of examples; and the frequency annotation based on the available number of examples for the given species within the dataset.

\begin{table}[ht]
\centering
\caption{Recognized tree species and common names for the given genera.}
\label{tab:genus-species}
\scriptsize
\begin{tabular}{lp{2in}p{2in}}
\toprule
{\bf Genus} &                                  {\bf Tree pecies} &                                {\bf Common names} \\
\midrule
Abies       &                                Abies sachalinensis &                                       Sakhalin Fir \\
Acacia      &                                 Acacia melanoxylon &                Acacia/Wattle, Australian Blackwood \\
Acer        &                                        Acer pictum &                                         Mono Maple \\
Alnus       &                                                    &                                              Alder \\
Anacardium  &                             Anacardium occidentale &                                             Cashew \\
Araucaria   &                                                    &                                      Monkey Puzzel \\
Areca       &                                                    &                                         Areca Palm \\
Betula      &                                     Betula pendula &                             East Asian White Birch \\
Callitris   &                                                    &                                       Cypress Pine \\
Camellia    &                                      Thea sinensis &                                                Tea \\
Castanea    &                                   Castanea crenata &                                    Korean Chestnut \\
Casuarina   &                                                    &                                          Casuarina \\
Cedrus      &                                                    &                                              Cedar \\
Citrus      &                                                    &                                             Orange \\
Cocos       &                                     Cocos nucifera &                                       Coconut Palm \\
Coffea      &                                                    &                                             Coffee \\
Cornus      &                                 Cornus controversa &                                       Wedding Cake \\
Cryptomeria &                               Cryptomeria japonica &                                 Japanese Red Cedar \\
Dendropanax &                             Dendropanax morbiferus &                                 Korean Dendropanax \\
Elaeis      &                                  Elaeis guineensis &                                           Oil Palm \\
Eucalyptus  &             Eucalyptus globulus, Eucalyptus nitens &         Eucalyptus, Tasmanian Bluegum, Shining Gum \\
Fraxinus    &                             Fraxinus rhynchophylla &                                     East Asian Ash \\
Ginkgo      &                                      Ginkgo biloba &                                             Ginkgo \\
Gliricidia  &                                                    &                                         Gliricidia \\
Grevillea   &                                                    &                                          Grevillea \\
Hevea       &                                 Hevea brasiliensis &                                             Rubber \\
Jacaranda   &                                                    &                                          Jacaranda \\
Larix       &                                                    &                                              Larch \\
Lithocarpus &                                     Pasania edulis &                                 Japanese Stone Oak \\
Machilus    &                                Machilus thunbergii &                                  Japanese Bay Tree \\
Malus       &                                       Malus pumila &                                              Apple \\
Mangifera   &                                                    &                                              Mango \\
Morus       &                                                    &                                           Mulberry \\
Musa        &                                                    &                                             Banana \\
Picea       &      Picea glehnii, Picea jezoensis, Picea mariana &  Spruce, Sakhalin Spruce, Jezo Spruce, Black Spruce \\
Pinus       &  Pinus densiflora, Pinus thunbergii, Pinus koraiensis, Pinus rigida, Pinus elliottii, Pinus taeda, Pinus echinata, Pinus radiata &  Red Pine, Black Pine, Korean Pine, Pitch Pine, Slash Pine, Loblolly Pine, Shortleaf Pine, Pine, Monterey Pine \\
Populus     &                                                    &                                             Poplar \\
Prunus      &                    Prunus serrulata, Prunus dulcis &                  Japanese Flowering Cherry, Almond \\
Pseudotsuga &                              Pseudotsuga menziesii &                                        Douglas Fir \\
Pterocarpus &                                                    &                                             Padauk \\
Quercus     &             Quercus acutissima, Quercus variabilis &                Oak, Sawtooth Oak, Chinese Cork Oak \\
Robinia     &                               Robinia pseudoacacia &                                       Black Locust \\
Shorea      &                                     Shorea robusta &                                                Sal \\
Tectona     &                                                    &                                               Teak \\
Theobroma   &                                    Theobroma cacao &                                              Cacao \\
Zelkova     &                                    Zelkova serrata &                                    Sawleaf Zelkova \\
\bottomrule
\end{tabular}
\end{table}

\begin{table}[ht]
\centering
\caption{Distribution of genus labels across training splits with frequency annotation (columns: genus, total number of examples, train/validation/test examples, frequency annotation of the given genera within the dataset).}
\label{tab:labels-distribution-genera}
\scriptsize
\begin{tabular}{rlrrrrll}
\toprule
& {\bf Tree genus} & {\bf Total} & {\bf Train} & {\bf Validation} & {\bf Test} & {\bf Frequency}\\
\midrule
1 & Pinus & 929381 (41.037\%) & 735144 & 94916 & 99321 & frequent \\
2 & Elaeis & 466895 (20.616\%) & 369606 & 49603 & 47686 & frequent \\
3 & Eucalyptus & 271718 (11.998\%) & 212349 & 27706 & 31663 & frequent \\
4 & Larix & 196918 (8.695\%) & 147690 & 29215 & 20013 & frequent \\
5 & Abies & 57864 (2.555\%) & 40703 & 8982 & 8179 & frequent \\
6 & Tectona & 43840 (1.936\%) & 26278 & 11846 & 5716 & frequent \\
7 & Shorea & 42267 (1.866\%) & 28532 & 7772 & 5963 & frequent \\
8 & Castanea & 38537 (1.702\%) & 20344 & 9964 & 8229 & frequent \\
9 & Pseudotsuga & 36350 (1.605\%) & 28704 & 3690 & 3956 & frequent \\
10 & Populus & 25159 (1.111\%) & 18518 & 3052 & 3589 & frequent \\
11 & Acacia & 24885 (1.099\%) & 16645 & 5740 & 2500 & frequent \\
12 & Morus & 19921 (0.880\%) & 13717 & 4093 & 2111 & frequent \\
13 & Robinia & 19912 (0.879\%) & 14319 & 3400 & 2193 & frequent \\
14 & Picea & 17893 (0.790\%) & 13469 & 1878 & 2546 & frequent \\
15 & Areca & 11853 (0.523\%) & 8385 & 2930 & 538 & frequent \\
16 & Camellia & 11301 (0.499\%) & 7477 & 1153 & 2671 & frequent \\
17 & Betula & 10387 (0.459\%) & 7305 & 2043 & 1039 & frequent \\
18 & Quercus & 9904 (0.437\%) & 6790 & 2053 & 1061 & common \\
19 & Hevea & 8450 (0.373\%) & 6506 & 848 & 1096 & common \\
20 & Cedrus & 7357 (0.325\%) & 4822 & 1530 & 1005 & common \\
21 & Mangifera & 2859 (0.126\%) & 1323 & 294 & 1242 & common \\
22 & Cocos & 1964 (0.087\%) & 1216 & 204 & 544 & common \\
23 & Anacardium & 1544 (0.068\%) & 772 & 498 & 274 & common \\
24 & Prunus & 1287 (0.057\%) & 810 & 217 & 260 & common \\
25 & Zelkova & 1098 (0.048\%) & 840 & 132 & 126 & common \\
26 & Alnus & 862 (0.038\%) & 561 & 99 & 202 & common \\
27 & Acer & 815 (0.036\%) & 630 & 90 & 95 & common \\
28 & Coffea & 738 (0.033\%) & 420 & 80 & 238 & common \\
29 & Fraxinus & 433 (0.019\%) & 314 & 44 & 75 & common \\
30 & Musa & 364 (0.016\%) & 245 & 48 & 71 & common \\
31 & Casuarina & 297 (0.013\%) & 223 & 36 & 38 & common \\
32 & Ginkgo & 290 (0.013\%) & 190 & 54 & 46 & common \\
33 & Callitris & 257 (0.011\%) & 166 & 39 & 52 & common \\
34 & Grevillea & 229 (0.010\%) & 156 & 36 & 37 & common \\
35 & Lithocarpus & 197 (0.009\%) & 122 & 37 & 38 & rare \\
36 & Dendropanax & 136 (0.006\%) & 67 & 35 & 34 & rare \\
37 & Gliricidia & 134 (0.006\%) & 46 & 36 & 52 & rare \\
38 & Machilus & 124 (0.005\%) & 56 & 34 & 34 & rare \\
39 & Cryptomeria & 112 (0.005\%) & 66 & 37 & 9 & rare \\
40 & Malus & 52 (0.002\%) & 26 & 12 & 14 & rare \\
41 & Pterocarpus & 39 (0.002\%) & 14 & 18 & 7 & rare \\
42 & Citrus & 38 (0.002\%) & 12 & 14 & 12 & rare \\
43 & Jacaranda & 31 (0.001\%) & 9 & 13 & 9 & rare \\
44 & Cornus & 25 (0.001\%) & 8 & 7 & 10 & rare \\
45 & Araucaria & 15 (0.001\%) & 5 & 5 & 5 & rare \\
46 & Theobroma & 15 (0.001\%) & 6 & 4 & 5 & rare \\
\bottomrule
\end{tabular}
\end{table}

\begin{table}[ht]
\centering
\caption{Distribution of tree species labels across training splits with frequency annotation (columns: tree species, total number of examples, train/validation/test examples, frequency annotation of the given species within the dataset).}
\label{tab:labels-distribution}
\scriptsize
\begin{tabular}{rlrrrrll}
\toprule
& {\bf Tree species} & {\bf Total} & {\bf Train} & {\bf Validation} & {\bf Test} & {\bf Frequency}\\
\midrule
1 & Elaeis guineensis & 466895 (34.631\%) & 373074 & 47103 & 46718 & frequent \\
2 & Pinus taeda & 205606 (15.250\%) & 164474 & 20571 & 20561 & frequent \\
3 & Pinus rigida & 125575 (9.314\%) & 100230 & 12719 & 12626 & frequent \\
4 & Pinus koraiensis & 99082 (7.349\%) & 79076 & 9969 & 10037 & frequent \\
5 & Pinus densiflora & 67861 (5.033\%) & 54167 & 6811 & 6883 & frequent \\
6 & Abies sachalinensis & 56038 (4.156\%) & 44690 & 5669 & 5679 & frequent \\
7 & Pinus elliottii & 43072 (3.195\%) & 34406 & 4318 & 4348 & frequent \\
8 & Shorea robusta & 42267 (3.135\%) & 33319 & 4408 & 4540 & frequent \\
9 & Castanea crenata & 38537 (2.858\%) & 30569 & 4099 & 3869 & frequent \\
10 & Pseudotsuga menziesii & 36350 (2.696\%) & 29059 & 3641 & 3650 & frequent \\
11 & Eucalyptus globulus & 29643 (2.199\%) & 23661 & 2966 & 3016 & frequent \\
12 & Pinus radiata & 24963 (1.852\%) & 19930 & 2516 & 2517 & frequent \\
13 & Robinia pseudoacacia & 19912 (1.477\%) & 15743 & 2168 & 2001 & frequent \\
14 & Pinus thunbergii & 18478 (1.371\%) & 14735 & 1893 & 1850 & frequent \\
15 & Pinus echinata & 13532 (1.004\%) & 10803 & 1356 & 1373 & frequent \\
16 & Thea sinensis & 11301 (0.838\%) & 8983 & 1155 & 1163 & frequent \\
17 & Betula pendula & 10387 (0.770\%) & 8297 & 1043 & 1047 & frequent \\
\midrule
18 & Hevea brasiliensis & 8450 (0.627\%) & 6683 & 916 & 851 & common \\
19 & Picea glehnii & 8263 (0.613\%) & 6589 & 840 & 834 & common \\
20 & Quercus acutissima & 6961 (0.516\%) & 5551 & 713 & 697 & common \\
21 & Eucalyptus nitens & 5630 (0.418\%) & 4477 & 571 & 582 & common \\
22 & Cocos nucifera & 1964 (0.146\%) & 1555 & 200 & 209 & common \\
23 & Anacardium occidentale & 1544 (0.115\%) & 1208 & 178 & 158 & common \\
24 & Zelkova serrata & 1098 (0.081\%) & 878 & 110 & 110 & common \\
25 & Acer pictum & 815 (0.060\%) & 647 & 82 & 86 & common \\
26 & Prunus serrulata & 653 (0.048\%) & 519 & 66 & 68 & common \\
27 & Prunus dulcis & 634 (0.047\%) & 486 & 76 & 72 & common \\
28 & Picea jezoensis & 622 (0.046\%) & 491 & 68 & 63 & common \\
29 & Quercus variabilis & 597 (0.044\%) & 467 & 64 & 66 & common \\
30 & Fraxinus rhynchophylla & 433 (0.032\%) & 345 & 44 & 44 & common \\
31 & Ginkgo biloba & 290 (0.022\%) & 218 & 34 & 38 & common \\
\midrule
32 & Pasania edulis & 197 (0.015\%) & 122 & 35 & 40 & rare \\
33 & Dendropanax morbiferus & 136 (0.010\%) & 67 & 34 & 35 & rare \\
34 & Machilus thunbergii & 124 (0.009\%) & 58 & 36 & 30 & rare \\
35 & Cryptomeria japonica & 112 (0.008\%) & 50 & 28 & 34 & rare \\
36 & Acacia melanoxylon & 54 (0.004\%) & 24 & 13 & 17 & rare \\
37 & Malus pumila & 52 (0.004\%) & 17 & 9 & 26 & rare \\
38 & Picea mariana & 51 (0.004\%) & 17 & 22 & 12 & rare \\
39 & Cornus controversa & 25 (0.002\%) & 10 & 8 & 7 & rare \\
40 & Theobroma cacao & 15 (0.001\%) & 6 & 5 & 4 & rare \\
\bottomrule
\end{tabular}
\end{table}

\begin{table}[ht]
\centering
\caption{Planted trees traits (columns: common species name; official species name; whether it is a timber or tree crops plantation; whether those are conifer or broad leaf trees; whether evergreen or deciduous; and whether those trees are considered to have soft or hard wood).}
\label{tab:traits}
\scriptsize
\begin{tabular}{rllllll}
\toprule
& {\bf Common name} & {\bf Species/genus name} & {\bf timber / crops} &  {\bf confr / broadl} & {\bf ever / dec} & {\bf soft / hard}\\
\midrule
1 & Oil Palm & Elaeis guineensis & Tree crops & Unknown & Unknown & Unknown \\
2 & Pine & Pinus sp. & Planted forest & Conifer & Evergreen & Softwood \\
3 & Eucalyptus & Eucalyptus sp. & Planted forest & Broadleaf & Evergreen & Hardwood \\
4 & Loblolly Pine & Pinus taeda & Planted forest & Conifer & Evergreen & Softwood \\
5 & Larch & Larix sp. & Planted forest & Conifer & Deciduous & Softwood \\
6 & Pitch Pine & Pinus rigida & Planted forest & Conifer & Evergreen & Softwood \\
7 & Korean Pine & Pinus koraiensis & Planted forest & Broadleaf & Evergreen & Softwood \\
8 & Red Pine & Pinus densiflora & Planted forest & Conifer & Evergreen & Softwood \\
9 & Sakhalin Fir & Abies sachalinensis & Planted forest & Conifer & Evergreen & Softwood \\
10 & Teak & Tectona sp. & Planted forest & Broadleaf & Deciduous & Hardwood \\
11 & Slash Pine & Pinus elliottii & Planted forest & Conifer & Evergreen & Softwood \\
12 & Sal & Shorea robusta & Planted forest & Broadleaf & Evergr/Decid & Hardwood \\
13 & Korean Chestnut & Castanea crenata & Tree crops & Unknown & Unknown & Unknown \\
14 & Douglas Fir & Pseudotsuga menziesii & Planted forest & Conifer & Evergreen & Softwood \\
15 & Tasmanian Bluegum & Eucalyptus globulus & Planted forest & Broadleaf & Evergreen & Hardwood \\
16 & Poplar & Populus sp. & Planted forest & Broadleaf & Deciduous & Hardwood \\
17 & Monterey Pine & Pinus radiata & Planted forest & Conifer & Evergreen & Softwood \\
18 & Acacia/Wattle & Acacia sp. & Planted forest & Broadleaf & Deciduous & Hardwood \\
19 & Mulberry & Morus sp. & Tree crops & Unknown & Unknown & Unknown \\
20 & Black Locust & Robinia pseudoacacia & Planted forest & Broadleaf & Deciduous & Hardwood \\
21 & Black Pine & Pinus thunbergii & Planted forest & Conifer & Evergreen & Softwood \\
22 & Shortleaf Pine & Pinus echinata & Planted forest & Conifer & Evergreen & Softwood \\
23 & Areca Palm & Areca sp. & Tree crops & Unknown & Unknown & Unknown \\
24 & Tea & Thea sinensis & Tree crops & Unknown & Unknown & Unknown \\
25 & East Asian White Birch  & Betula pendula & Planted forest & Broadleaf & Deciduous & Unknown \\
26 & Spruce & Picea sp. & Planted forest & Conifer & Evergreen & Softwood \\
27 & Rubber & Hevea brasiliensis & Tree crops & Unknown & Unknown & Unknown \\
28 & Sakhalin Spruce & Picea glehnii & Planted forest & Conifer & Evergreen & Softwood \\
29 & Cedar & Cedrus sp. & Planted forest & Conifer & Evergreen & Softwood \\
30 & Sawtooth Oak & Quercus acutissima & Planted forest & Broadleaf & Deciduous & Hardwood \\
31 & Shining Gum & Eucalyptus nitens & Planted forest & Broadleaf & Evergreen & Hardwood \\
32 & Mango & Mangifera sp. & Tree crops & Unknown & Unknown & Unknown \\
33 & Oak & Quercus sp. & Planted forest & Broadleaf & Unknown & Hardwood \\
34 & Coconut Palm & Cocos nucifera & Tree crops & Unknown & Unknown & Unknown \\
35 & Fir & Abies sp. & Planted forest & Conifer & Evergreen & Softwood \\
36 & Cashew & Anacardium occidentale & Tree crops & Unknown & Unknown & Unknown \\
37 & Sawleaf Zelkova & Zelkova serrata & Planted forest & Broadleaf & Deciduous & Hardwood \\
38 & Alder & Alnus sp. & Planted forest & Broadleaf & Deciduous & Hardwood \\
39 & Mono Maple & Acer pictum & Planted forest & Broadleaf & Deciduous & Hardwood \\
40 & Coffee & Coffea sp. & Tree crops & Unknown & Unknown & Unknown \\
41 & Japanese Flowering Cherry & Prunus serrulata & Tree crops & Broadleaf & Deciduous & Hardwood \\
42 & Almond & Prunus dulcis & Tree crops & Unknown & Unknown & Unknown \\
43 & Jezo Spruce & Picea jezoensis & Planted forest & Conifer & Evergreen & Softwood \\
44 & Chinese Cork Oak & Quercus variabilis & Tree crops & Broadleaf & Deciduous & Unknown \\
45 & East Asian Ash & Fraxinus rhynchophylla & Planted forest & Broadleaf & Deciduous & Hardwood \\
46 & Banana & Musa sp. & Tree crops & Unknown & Unknown & Unknown \\
47 & Casuarina & Casuarina sp. & Planted forest & Broadleaf & Evergreen & Hardwood \\
48 & Ginkgo & Ginkgo biloba & Tree crops & Unknown & Unknown & Unknown \\
49 & Cypress Pine & Callitris sp. & Planted forest & Conifer & Evergreen & Softwood \\
50 & Grevillea & Grevillea sp. & Planted forest & Broadleaf & Evergreen & Hardwood \\
51 & Japanese Stone Oak & Pasania edulis & Tree crops & Broadleaf & Evergreen & Unknown \\
52 & Korean Dendropanax & Dendropanax morbiferus & Tree crops & Unknown & Unknown & Unknown \\
53 & Gliricidia & Gliricidia sp. & Tree crops & Unknown & Unknown & Unknown \\
54 & Japanese Bay Tree & Machilus thunbergii & Planted forest & Broadleaf & Evergreen & Hardwood \\
55 & Japanese Red Cedar & Cryptomeria japonica & Planted forest & Conifer & Evergreen & Softwood \\
56 & Australian Blackwood & Acacia melanoxylon & Planted forest & Broadleaf & Evergreen & Hardwood \\
57 & Apple & Malus pumila & Tree crops & Unknown & Unknown & Unknown \\
58 & Black Spruce & Picea mariana & Planted forest & Conifer & Evergreen & Softwood \\
59 & Padauk & Pterocarpus sp. & Planted forest & Broadleaf & Evergr/Decid & Hardwood \\
60 & Orange & Citrus sp. & Tree crops & Unknown & Unknown & Unknown \\
61 & Jacaranda & Jacaranda sp. & Planted forest & Broadleaf & Deciduous & Hardwood \\
62 & Wedding Cake & Cornus controversa  & Tree crops & Broadleaf & Deciduous & Unknown \\
63 & Cacao & Theobroma cacao & Tree crops & Unknown & Unknown & Unknown \\
64 & Monkey Puzzel & Araucaria sp. & Planted forest & Conifer & Evergreen & Softwood \\
\bottomrule
\end{tabular}
\end{table}

Each of the 64 label categories has a unique species name and common name. Next, we can assign each species to either timber or tree crop class, to being conifer or broad-leaf trees, to be evergreen or deciduous (some species can be both, in dependence of climate conditions), and to have soft or hard wood. \autoref{tab:traits} lists all correspondences. If a specific value was not provided, we mark it as \textit{Unknown}.

\clearpage

\section{Model and training details}
\label{app:details-model}
We used a standard vision transformer (ViT) architecture \cite{dosovitskiy2021:vit} for single-modality results. The default configuration consisted of 12 transformer layers, each consisting of a multi-head (3) self-attention blocks and an MLP block with size 768, with layer normalization. The used embedding size is 192. Further model, optimizer and training configuration parametrs are shown in \autoref{tab:model-config}.
For multi-temporal data, we extended the patching from spatial 2D to spatio-temporal 3D patches. For multi-modal experiments, to evaluate early, mid-, and late fusion, we followed \cite{nagrani2021attention}. We considered four training data augmentation approaches: none, rotation and flipping, temporal masking, and combination the last two.

\begin{table}
\centering
\caption{Default model and training configuration.}
\label{tab:model-config}
\begin{tabular}{lr}
\toprule
Name & Value \\
\midrule
transformer layers & 12 \\
embedding size & 192 \\
number of heads & 3 \\
MLP size & 768 \\
optimizer & adamw \\
base learning rate & 0.001 \\
weight decay & 0.0001 \\
batch size & 1024 \\
training epochs & 40 \\
learning rte schedule & cosine decay \\
warmup & 5 epochs \\
loss & softmax cross-entropy \\
\bottomrule
\end{tabular}
\end{table}

\section{Additional experimental results}
\label{app:details-exps}

\begin{figure}[htbp]
    \centering
    \includegraphics[width=1.0\linewidth]{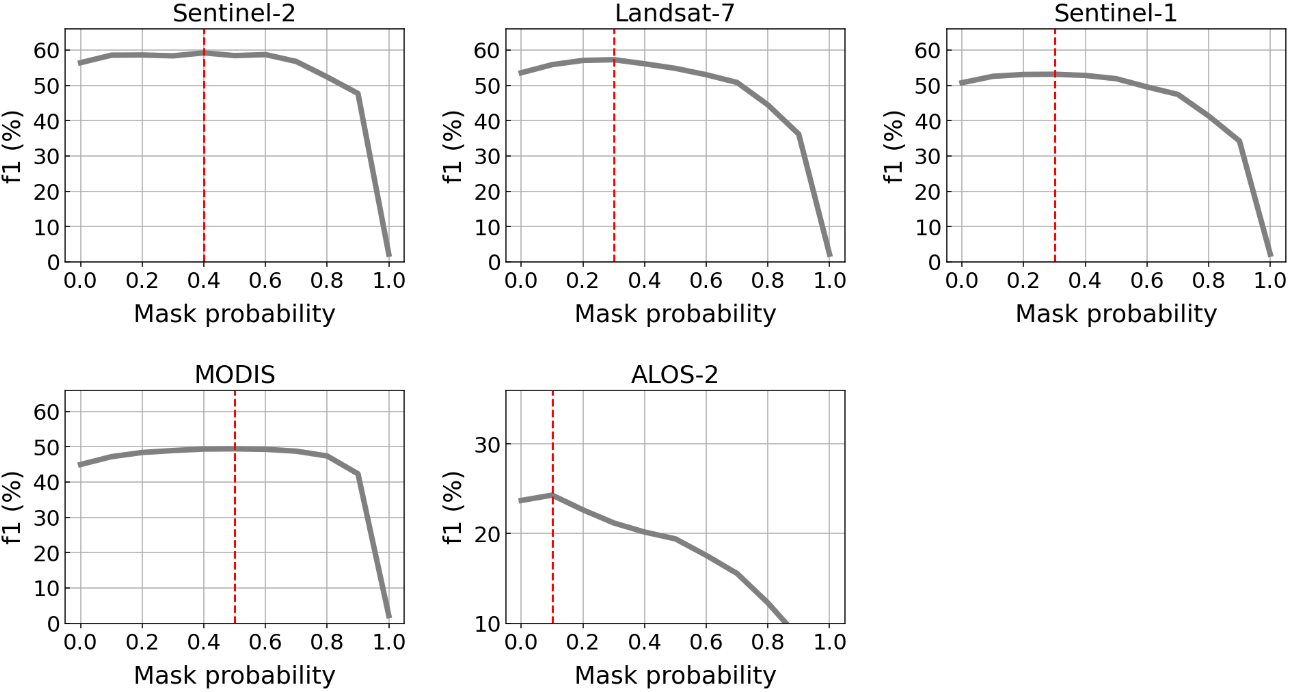}
    \caption{Validation F1 score over temporal masking probability for individual satellites.}
    \label{fig:mask-ratios}
\end{figure}

\autoref{fig:mask-ratios} visualizes an ablation across different temporal masking probabilities for all individual satellite modalities. As can be observed, the best temporal masking probability for most satellites is between 30\% and 50\%, except for Alos data, where the best is at 10\%. This result is intuitive, since Alos data has the fewest temporal samples and therefore relies a lot on all given data.

\autoref{tab:baselines} presents reference results using Random Forests or MLP models, where AutoML was used to optimize the configurations.

\begin{table}
\centering
\caption{Baseline F1 metrics for random forests and neural networks. ({\bf F1${}_r$},  
{\bf F1${}_c$} and {\bf F1${}_f$} denote
\textit{rare}, \textit{common} and \textit{frequent} classes
respectively). Optimal models were selected based on validation accuracy. }
\label{tab:baselines}
\resizebox{\textwidth}{!}{
\large
\begin{tabular}{lc@{\hskip 0.15in}cccc@{\hskip 0.1in}cccc}
\toprule
{\bf Satellites} & {\bf Model} &
 \multicolumn{4}{c}{\bf Validation}& \multicolumn{4}{c}{\bf Test}
\\
& &{\bf F1 Macro}&{\bf F1${}_r$}&{\bf F1${}_c$}&{\bf F1${}_f$} 
&{\bf F1 Macro}&{\bf F1${}_r$}&{\bf F1${}_c$}&{\bf F1${}_f$} 
\\
\cmidrule(r){3-6}
\cmidrule(r){7-10}
s2& NN & 49.5 & 17.1 & 34.1 & 79.4 & 50.6 & 20.4 & 34.7 & 79.7\\
s2& RF & 30.7 & 1.5 & 9.1 & 63.9 & 32.0 & 4.2 & 11.5 & 63.5\\
s2-s1-modis-l7-alos& NN & 51.0 & 22.3 & 34.5 & 79.9 & 51.8 & 26.2 & 34.0 & 80.1\\
s2-s1-modis-l7-alos& RF & 39.7 & 5.4 & 24.9 & 70.0 & 41.6 & 7.9 & 25.4 & 72.8\\
\bottomrule
\end{tabular}
}
\end{table}

\end{document}